\RequirePackage{amsthm}

\documentclass[pdflatex, iicol, sn-mathphys-num]{sn-jnl}

\usepackage{graphicx}%
\usepackage{multirow}%
\usepackage{amsmath,amssymb,amsfonts}%
\usepackage{mathrsfs}%
\usepackage[title]{appendix}%
\usepackage{xcolor}%
\usepackage{textcomp}%
\usepackage{manyfoot}%
\usepackage{booktabs}%
\usepackage{algorithm}%
\usepackage{algorithmicx}%
\usepackage{algpseudocode}%
\usepackage{listings}%
\usepackage{amssymb}
\usepackage{siunitx}
\usepackage{dirtytalk}
\usepackage{footnote}
\usepackage{float}
\usepackage{hyperref}
\usepackage{booktabs}
\usepackage[T1]{fontenc}
\usepackage{graphicx}
\usepackage{tabularx}
\usepackage{multirow}
\usepackage{amsmath}
\usepackage[misc,geometry]{ifsym}
\usepackage{xcolor}
\usepackage{array} %
\newcolumntype{P}[1]{>{\centering\arraybackslash}p{#1}}
\newcolumntype{M}[1]{>{\centering\arraybackslash}m{#1}}

\NewDocumentCommand{\rot}{O{45} O{1em} m}{\makebox[#2][l]{\rotatebox{#1}{#3}}}%
\usepackage{ltablex}
\usepackage{graphicx}
\usepackage{cleveref}
\usepackage{kantlipsum}
\usepackage{multirow}
\usepackage{siunitx}
\usepackage{color,soul}
\usepackage{subcaption}
\usepackage{arydshln}
\usepackage[inline]{enumitem}
\crefname{enumi}{}{}
\usepackage[most]{tcolorbox}
\colorlet{Blue}{blue!40!}
\definecolor{RedClear}{rgb}{0.91,0.65,0.58}
\tcbset{on line, 
        boxsep=2pt, left=0pt,right=0pt,top=0pt,bottom=0pt,
        colframe=white,colback=RedClear        
        }

\theoremstyle{thmstyleone}%

\theoremstyle{thmstyletwo}%

\theoremstyle{thmstylethree}%

\begin{document}

\title[Explainable Depression Symptom Detection in Social Media]{Explainable Depression Symptom Detection in Social Media}

\author*[1]{\fnm{Eliseo} \sur{Bao}}\email{eliseo.bao@udc.es}
\equalcont{These authors contributed equally to this work.}

\author[1]{\fnm{Anxo} \sur{Pérez}}\email{anxo.pvila@udc.es}
\equalcont{These authors contributed equally to this work.}

\author[1]{\fnm{Javier} \sur{Parapar}}\email{javier.parapar@udc.es}
\equalcont{These authors contributed equally to this work.}

\affil*[1]{\orgdiv{Information Retrieval Lab (IRLab)}, \orgname{Centro de Investigación en Tecnoloxías da Información e da Comunicación (CITIC)}, \orgaddress{\street{Campus de Elviña}, \city{A Coruña}, \postcode{15071}, \state{Galicia}, \country{Spain}}}

\abstract{Users of social platforms often perceive these sites as supportive spaces to post about their mental health issues. Those conversations contain important traces about individuals' health risks. Recently, researchers have exploited this online information to construct mental health detection models, which aim to identify users at risk on platforms like Twitter, Reddit or Facebook. Most of these models are focused on achieving good classification results, ignoring the explainability and interpretability of the decisions. Recent research has pointed out the importance of using clinical markers, such as the use of symptoms, to improve trust in the computational models by health professionals. In this paper, we introduce transformer-based architectures designed to detect and explain the appearance of depressive symptom markers in user-generated content from social media. We present two approaches: $i)$ train a model to classify, and another one to explain the classifier's decision separately and $ii)$ unify the two tasks simultaneously within a single model. Additionally, for this latter manner, we also investigated the performance of recent conversational Large Language Models (LLMs) utilizing both in-context learning and finetuning. Our models provide natural language explanations, aligning with validated symptoms, thus enabling clinicians to interpret the decisions more effectively. We evaluate our approaches using recent symptom-focused datasets, using both offline metrics and expert-in-the-loop evaluations to assess the quality of our models’ explanations. Our findings demonstrate that it is possible to achieve good classification results while generating interpretable symptom-based explanations.}


\maketitle

\section{Introduction}

Mental health is a crucial component of overall health and well-being~\cite{PRINCE2007859}. The most recent figures estimate mental disorders prevalence in adults over 20\%~\cite{aus2022}. The World Health Organization (WHO) estimates that approximately 332 million people globally are affected by depression~\cite{world2017depression}. Early intervention in mental disorders is crucial in mitigating their impact, especially for young individuals~\cite{Picardi2016-jh}. However, due to the stigma associated with depression, more than 60\% of individuals with the condition do not seek professional support~\cite{Gulliver2010}. To address this problem, computational researchers have a growing interest in assisting in the early detection and diagnosis of depression, thereby mitigating its societal impact~\cite{de2013predicting}. 

In this scenario, researchers have found that the writings posted by individuals on social media platforms are valuable evidence for looking for early signs of depression~\cite{sadeque-measuring-depression-2018,yates-etal-2017-depression,cacheda2019early,behesti-mental-disorders-2020, bucur2023s,aragon-etal-2023-disorbert}. Individuals experiencing depression find comfort in expressing their thoughts and emotions on these platforms, motivated by factors such as privacy or anonymity~\cite{callahan2012cybertherapy,Ferraro2020}. Consequently, social media provides a complementary opportunity to access valuable information about individuals' state of mind beyond traditional professional therapy. The combined use of computational linguistic techniques and the vast amount of data from social networking has led to significant advancements in detecting signs of depression~\cite{rissola2021survey}. The field's critical nature motivated much effort in creating curated experimental benchmarks~\cite{parapar2023erisk,clpsych-2022-linguistics}, which allowed the development and evaluation of many new predictive models. Traditional efforts used engineered features such as word counts, posting activity or emotion levels~\cite{sadeque-measuring-depression-2018,yates-etal-2017-depression,cacheda2019early}. Recently, and due to the rise of transformer-based language models, many researchers used these deep learning models as classifiers to identify users at risk of depression and similar disorders in online environments~\cite{behesti-mental-disorders-2020, bucur2023s,aragon-etal-2023-disorbert}.

However, researchers in this field do not plan to replace mental health professionals but rather offer support to their work. Licensed clinicians play a crucial role in validating the predictions made by computational models and taking appropriate actions with individuals when necessary. Computational models may only be used carefully to extend those professionals' reach and facilitate their workflow. The existing models, however, present many limitations for achieving that goal~\cite{10.1093-jamiaopen-ooz054}. One significant barrier is their limited ability to explain their predictions. Reliable interpretation of models' decisions is mandatory for professionals to understand and trust these models and use them in their daily work~\cite{hauser2022promise}. One way to pursue that is by designing new models incorporating trustworthy and reliable explanations~\cite{Ernala-method-gap2019}. Following that path, recent research has explored the utilization of symptoms collected from validated clinical questionnaires. Most of these proposals, in the field of depression, used the markers from the Beck Depression Inventory-II (BDI-II)~\cite{Beck1996} or the 9-Question Patient Health Questionnaire (PHQ-9)~\cite{kroenke2001phq}, which encompass a range of depressive symptoms such as irritability, pessimism or sleep problems. The utilization of such symptom markers has been shown to improve the explainability, generalization and overall performance of depression detection models~\cite{PEREZ2022102380,nguyen-etal-2022-improving,zhang-etal-2022-symptom,ijcai2022p725,perez-etal-2023-semantic}.

With that motivation, we aim to develop models that categorise whether or not social media posts exhibit markers of validated depressive symptoms. Accurately detecting symptom information along a user's vast amount of writing is the first step in developing explainable depression detection models. We go beyond classifying posts for depressive symptoms by providing a explanation for the decisions. It is important to clarify that our paper aims not to diagnose depressive disorder, but rather to identify markers indicative of potential symptoms. The diagnosis of depressive disorder, which is the ultimately aim and scope of psychologists in this domain, relies on clinical factors such as the presence of clinical symptoms and their temporality~\cite{smith2013diagnosis}. For this reason, this work seeks to aid in this diagnostic process by generating explanations of possible depressive symptomatology. 

In this context, the terms `interpretability' and `explainability' can be challenging to define in our specific context, given the variations in their use across existing literature. Some authors use these terms interchangeably, referring to the ability to explain or present systems in a manner understandable to humans~\cite{doshivelez2017rigorous}. However, other authors consider them as distinct concepts. In this perspective, interpretability relates to the system's capacity to be understood by humans, while explainability encompasses being true to reality~\cite{MARKUS2021103655}. In our work, we adopt a simple definition: interpretability/explainability refers to the extent to which humans can understand the reasons behind a decision~\cite{MILLER20191}.

\begin{figure*}[ht]
	\centering
	\includegraphics[width=1\linewidth]{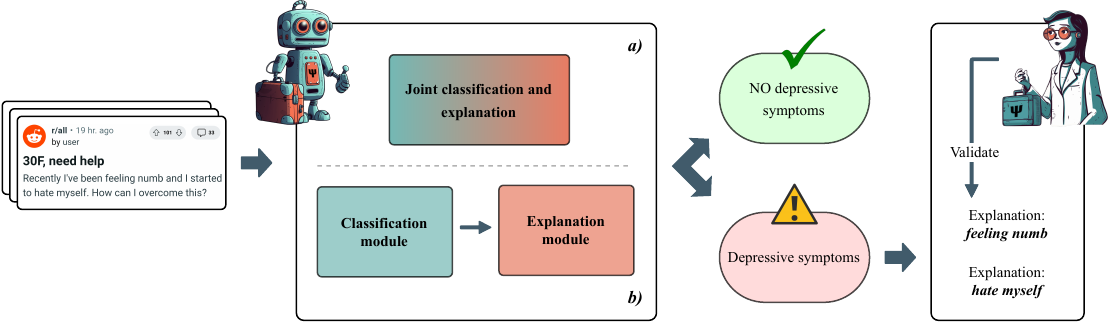}
	\caption{Overall pipeline of our proposals for the classification and generation of natural language explanations for the presence of depression symptom information in social media posts.}
	\label{fig:pipeline-overview}
\end{figure*}

In this paper, we introduce a text-to-text pipeline designed to achieve two main objectives: first, to classify the relevance of social media publications to depressive symptoms, and second, to provide explanations for these classifications. To implement this pipeline, we explore the effectiveness of state-of-the-art transformer-based models. Figure~\ref{fig:pipeline-overview} illustrates the two approaches considered: Part $a)$ employs text-to-text models that perform classification and explanation simultaneously. Part $b)$ corresponds with two-step approaches, utilizing separate models for classification and explanation. Additionally, we evaluate the capabilities of Large Language Models (LLMs) using recent general purpose conversational models such as GPT-3.5 Turbo~\cite{chatgpt}, Vicuna-13B~\cite{vicuna2023} and Mistral-7B-Instruct~\cite{jiang2023mistral} within an few-shot in-context learning approach. We also investigate how finetuned conversational LLMs perform for this tasks using domain-specific models such as MentalLLaMA~\cite{yang2023mentalllama} and MentalMistral.

We utilize two datasets, PsySym~\cite{zhang-etal-2022-symptom} and BDI-Sen~\cite{perez2023sigir}, for model training and evaluation. These datasets consist of sentences from social media posts associated with depressive symptoms. As we will discuss later, these sentences can be used as explanations for a post to explain its relevance to a symptom. To assess the quality of the generated explanations, we employ a combination of offline and expert-in-the-loop metrics. The experimental results demonstrate the effectiveness of our methods. By prioritizing explainability, our approach bridges the gap between automated predictions and human understanding, facilitating more informed clinical decision-making\footnote{Our implementation is available at: \url{https://gitlab.irlab.org/irlab/explainable-depression-symptom-detection-social-media}}.

Our study aims to address the following research questions:
\begin{enumerate*}[label=\textbf{RQ\arabic*}]
	\item \label{rq1}
	Can we train transformer-based models to accurately classify the presence of depressive symptoms in social media posts, while providing explanations for their decisions?
	
	\item \label{rq2}
	How many training examples, i.e. hand-labelled posts by domain experts, are needed to generate good explanations?
	
	\item \label{rq3}
	Does a unified model, designed for both classification and explanation, significantly differ in performance from using two separated models trained for each task?
	
	\item \label{rq4}
	How do recent conversational LLMs perform in detecting and explaining the presence of depressive symptoms in social media content?
\end{enumerate*}


\section{Related work}

Social media data has gained increasing attention for developing depression detection models in recent years~\cite{Islam2018, Li2023, GUNTUKU201743, 10.1145/3422824,crestani2022early}. The evaluation results of those methods show promising accuracy numbers~\cite{rissola2021survey,behesti-mental-disorders-2020, bucur2023s,tao-2017-eating-disorders,aragon-etal-2023-disorbert}. Such platforms offer an opportunity to identify disorders at an early stage~\cite{10.1145/2872427.2882996}, a step that is crucial to reduce negative impacts and their associated costs~\cite{Halfin2007-ea, Picardi2016-jh}. 

However, depression detection field still needs to overcome the cross-cutting problem of many machine learning applications: explainability. There are two common approaches when explaining models' decisions. On the one hand, we may use intrinsically explainable models, such as decision trees or linear models~\cite{8852158}. On the other hand, we can opt for model-agnostic, typically post hoc, explainability methods~\cite{2016arXiv160605386T, Ribeiro_Singh_Guestrin_2018}, which can be applied to any supervised machine learning model, regardless of its architecture. Moreover, the explanations can be local or global. Local explanations focus on individual decisions, allowing users to understand why the model produces a particular decision~\cite{10.5555/1756006.1859912, ribeiro2016should}. These local explanations, such as LIME (Local Interpretable Model-Agnostic Explanations)~\cite{ribeiro2016should} and SHAP (Shapley Additive exPlanations)~\cite{lundberg2017unified}, have already been used in the social psychology field~\cite{doi:10.1177/08944393221084064, NORDIN2023103316}. Alternatively, global explanations consider the whole machine learning model behaviour and its predictions altogether~\cite{Ustun2016, 10.1145/2939672.2939874, 10.5555/3295222.3295230}. Most traditional detection methods used engineered features (e.g., word counts or sentiment analysis) and based their global explanations on feature importance metrics~\cite{cacheda2019early,rissola2021survey}. These measures quantify the impact of each feature on the predictions, providing insights into the relative importance of each one.

With the rise of transformer-based models~\cite{NIPS2017_3f5ee243}, researchers have leveraged these models for sentence classification tasks to detect signs of depression in social media users, achieving impressive results in experimental benchmarks~\cite{parapar2023erisk, clpsych-2022-linguistics}. The encoder-decoder architecture of transformers effectively captures contextual information from input sequences and generates corresponding output sequences. However, maintaining the contextual integrity of each token in extensive texts presents a challenge. The attention mechanism tackles this by allocating weights to tokens relative to all others, allowing the model to focus on relevant parts of the input. Prior works have explored the attention mechanism's as an explainability tool~\cite{mullenbach-etal-2018-explainable, song-etal-2018-feature, 10.1007/978-3-030-51310-8_21}. Despite its utility, attention mechanisms also have limitations, particularly in interpreting the weights assigned to input features. There is an ongoing discussion in the community about the utility of attention weights for interpretability purposes, with some research advocating its benefits~\cite{wiegreffe-pinter-2019-attention, 9762294}, while others point out its constraints~\cite{jain-wallace-2019-attention, serrano-smith-2019-attention}. 

More recently, an alternative strategy for building interpretable models has emerged: the use of generative natural language explanations. This technique has several benefits: $i)$ they are readily comprehensible to end users, $ii)$ human annotators can more easily work with natural language, simplifying data collection, and $iii)$ it may be feasible to extract natural language explanations from large datasets of domain-expert data, a promising prospect for future research~\cite{NEURIPS2018_4c7a167b}. 

Linked to the recent breakthrough in the field with the advent of modern LLMs,~\citet{yang2023interpretable} explored their application in zero-shot/few-shot mental health analysis, assessing the influence of various emotion-enhanced prompts. This study investigates the potential of LLMs for explainable mental health assessment, elucidating the predictions through Chain-of-Thought (CoT) prompting~\cite{NEURIPS2022_8bb0d291}. While the study reveals that ChatGPT is capable of generating explanations at a human-level, it also notes that the approach has limitations, including unstable predictions and inaccurate reasoning. Following this study,~\citet{yang2023mentalllama} also introduced an interpretable mental health instruction dataset, constructed with ChatGPT-generated explanations. The authors used this data to train MentalLLaMA, a new open-source LLM that focuses on interpretable mental health analysis with conversational and reasoning skills. MentalLLaMA covers not only depression but also other related mental health conditions such as stress or suicide. In contrast, our study adopts a symptom-based approach, focusing on clinical markers for a nuanced analysis of depression. We include MentalLLaMA in our evaluations as a baseline, examining its performance against general-purpose models and its efficacy in symptom-level analysis. Our proposed method, MentalMistral, is another LLM fine-tuned for depression analysis but specifically at the symptom level. Notably, MentalMistral differs from MentalLLaMA in its data and explanation generation approach. While MentalLLaMA generates free-form explanations, MentalMistral is fine-tuned to generate explanations derived directly from segments of the input text, ensuring that its insights are firmly rooted in the original user-generated and human-annotated content. This approach ensures that our explanations are grounded in the text and provides a direct link between the user's words and the model's output.

Our goal is to cross the bridge by following prior efforts in training text-to-text models in order to produce natural language explanations for the depression detection problem by grounding all the explanations in clinical symptoms following the BDI-II questionnaire. As the explanations are formulated in human-readable text, our method makes the model's rationale behind the predictions transparent to clinicians. This level of interpretability is of critical importance to making informed decisions for depression detection~\cite{10.1093-jamiaopen-ooz054}, promoting both the efficacy and the trustworthiness of our models.


\section{Our proposal}

In this section, we describe our proposal for generating explainable decisions in the context of detecting depressive symptoms in social media posts. The task is defined as follows: given a user post, the model classifies whether it is indicative (positive) or non-indicative (negative) of any depressive symptom. A positive classification implies that the post reflects information for the user on one or more of the 21 symptoms described in the BDI-II questionnaire. Moreover, our objective extends to providing natural language explanations for any positive classification, showing the reasoning behind the model's decision. We undertake this task in two different manners: either as a single output (i.e. train together classification plus explanation), or in two separate steps (i.e. initial classification followed by post hoc explanation). These two scenarios are illustrated in Figure \ref{fig:pipeline-overview}. Part $a)$ is representative of the single-step approach, explained in Subsection~\ref{subsubsec:single_step}, while part $b)$ corresponds to the two-step approach, detailed in Subsection~\ref{subsubsec:two_stages}. If the pipeline identifies a post as having potential symptom risk, it provides corresponding explanations. Looking at the Figure, we can see that the explanations for the decision are \textit{`feeling numb'} and \textit{`hate myself'}. These generated explanations are designed to be reviewed and validated by health professionals.

\subsection{Explanations}
\label{subsec:notes-explanations}

Natural language explanation generation methods can be either extractive or abstractive~\cite{nle2023survey}. In the \emph{extractive} case, the model is asked to point out the parts of the original text that led to the decision. Conversely, \emph{abstractive explanations} involve the model to explain the reasons for the decision in a free format. Our approach focuses on extractive explanations, where the models highlights relevant text spans from the whole input text. This decision was driven by data availability and the advantage of this method in allowing quantitative analysis of explanation quality through overlap statistics with ground truth.

Furthermore, our models are designed to explain only the positive classifications. This decision comes from two considerations. First, explanations in negative cases (where no depressive symptom markers are detected) are less clinically relevant, as the primary focus is to identify positive indications of depression. Secondly, for extractive explanations, negatives cases would typically be a generic responses, such as \textit{"there is no evidence in the post about any depression symptom"}. Therefore, to optimize the utility and specificity of our models, we concentrate on explaining only the positive cases.

\subsection{Finetuning text-to-text Models}
\label{subsec:finetuning_text_to_text_models} 

\citet{10.5555/3455716.3455856} proposed the idea of transforming text-related tasks to follow the sequence-to-sequence (seq2seq) format~\cite{10.5555/2969033.2969173}, which is referred to as the text-to-text framework. This diverges from conventional methods like BERT-based approaches, which train models to yield a probability distribution over predefined output classes~\cite{devlin-etal-2019-bert}. In contrast, text-to-text models are trained to generate textual sequences. Consequently, these models may generate an unexpected output, which is considered a prediction error. 

In this context, the input text sequence, which we will call $\tilde{x}$, is defined as \small\texttt{``<task\_prefix>: <input\_text>''}\normalsize, where \small\texttt{<task\_prefix> }\normalsize identifies the task to be performed by the model. On the other hand, the output text sequence $\tilde{y}$ will be of the form \small\texttt{``<target>''}\normalsize, where \small\texttt{<target> }\normalsize corresponds to the desired output. If we take a sentiment analysis task as example, the sequence would be: $\tilde{x} = \small\texttt{``sentiment analysis: terrible product''}\normalsize$ and $\tilde{y} = \small\texttt{``negative''}\normalsize$. The Text-to-Text Transfer Transformer (T5) model is one of the most popular models under this paradigm~\cite{10.5555/3455716.3455856}, and it achieved state-of-the-art results in many NLP tasks~\cite{10.5555/3454287.3454581}. While there exist other text-to-text models, such as BART~\cite{lewis-etal-2020-bart}, prior research indicates that it performs worse than T5 in terms of explanation generation~\cite{10.1007/978-3-030-80599-9_8}. These models are pre-trained, providing us with a robust base that we can further finetune under either the single-step or two-steps approaches discussed below for generating explanations.

\subsubsection{Single-step}
\label{subsubsec:single_step}

\citet{narang2020wt5} explored how to teach text-to-text models to produce both classification and explanation. For that, they introduced an extension of T5 called WT5 ("Why T5?"). To use this method, the keyword {\small\texttt{``explain''}} is simply added to the input $\tilde{x}$, preceded by \small\texttt{``<task\_prefix>''}\normalsize. The target $\tilde{y}$ is appended with the phrase {\small\texttt{``explanation: <explanation>''}}. Using this template, we adapted it to our task resulting in the new input/output format:
{\small
	\begin{gather*} 
		\tilde{x} = \texttt{``explain symptom post:\;<post\_text>''}\\
		\tilde{y} = \texttt{``<target>\;[explanation:\;<explanation}_{1}\texttt{>]\;...}\\
		\texttt{...\;[explanation:\;<explanation}_{N}\texttt{>]''}
	\end{gather*}
}

The hard brackets denote potentially multiple explanation sentences. An example of input could be {\small\texttt{``explain symptom post: I absolutely hate myself (...) And I hate how I feel the need to burden other people with this. I am so whiny, so disgustingly insensitive (...)''}} \footnote{Full-length post is paraphrased and redacted for clarity and space reasons.}. Its corresponding target output would be  {\small\texttt{``positive explanation: I absolutely hate myself explanation: I am so whiny, so disgustingly insensitive''}}. Thus, in one step, users' posts can be both classified and explained in a extractive manner according to whether they indicate depressive symptoms or not. Part \textit{a)} in Figure \ref{fig:pipeline-overview} illustrates this approach, which we use to build two of our systems: \textbf{WT5} and \textbf{WBART}. Our WT5 system replicates the research of Narang et al. that we just briefly discussed, while our WBART system extends the same idea but to the BART (Bidirectional and Auto-Regressive Transformers) model~\cite{10.1007/978-3-030-80599-9_8}. 

While WT5 and WBART share the common objective of facilitating single-step classification and explanation, they diverge in their underlying architectures. WT5 builds upon the T5 model, whereas WBART utilizes the BART model. Both models adopt an encoder-decoder framework and are suitable for a wide range of sequence-to-sequence tasks. However, beyond considerations such as pretraining corpus, parameter initialization, and activation functions, the primary distinction between T5 and BART lies in their pretraining objectives. T5 implements a strategy where 15\% of tokens in the input sequence are randomly dropped out. These dropped-out tokens are replaced by sentinel tokens, and each one has a unique token ID assigned. The model's objective is to predict these dropped-out spans, delimited by the sentinel tokens. In contrast, BART's training process involves corrupting documents and optimizing a reconstruction loss between the decoder's output and the original document. Unlike T5, BART allows for various document corruptions, including the complete loss of source information.

\subsubsection{Two-steps pipeline}
\label{subsubsec:two_stages}

Alternatively, rather than training a single model to execute both classification and explanation, we propose a two-steps pipeline. In this approach, one model is initially employed for classification, followed by the use of a different model for explanation. We develop three of our systems using this methodology: \textbf{T5 + T5}, \textbf{BERT + T5}, and \textbf{MBERT + T5}. This process is depicted in part \textit{b)} of Figure \ref{fig:pipeline-overview}. Next, we define the stages of classification and explanation as they occur in this pipeline.

\paragraph{Classification} 

In alignment with recent studies, we conduct experiments with various types of pre-trained language models to investigate their capabilities in terms of classification~\cite{zhang-etal-2022-symptom, nguyen-etal-2022-improving,ijcai2022p725}. These models constitute the classification module and their sole responsibility is to determine whether a post is indicative of a depressive symptom. Firstly, we fine-tune a \textbf{T5} model with the aim of generating classification labels in a text format~\cite{10.5555/3455716.3455856}. Secondly, we explore the utility of BERT-based architectures, utilizing the pre-trained \textbf{BERT} base uncased model~\cite{devlin-etal-2019-bert}. Additionally, we finetuned MentalBERT (\textbf{MBERT}), a model purposely pre-trained for mental health applications using data compiled from various subreddits related to mental disorders~\cite{ji-etal-2022-mentalbert}.

\paragraph{Explanation}

For the explanation module, we finetuned a \textbf{T5} model responsible for explaining the evidence for those posts deemed positive by the classification models. As previously commented, we only trained this model to generate explanations for positive posts. Negative posts are discarded and do not enter in the second stage of the two-steps pipeline.

\subsection{In-context Learning with LLMs}
\label{subsec:in_context_learning}

In-context learning strategies allow guiding LLMs to specific behaviours within a given context without directly modifying their parameters~\cite{NEURIPS2020_1457c0d6}. In our study, we applied this technique to instruct recent conversational LLMs to perform our proposed single-step approach (\textit{classify + explain}). Using specific guidelines\footnote{See \ref{appendix:prompt} for details of the prompt strategy we adopted.}, we instructed the models to act as expert annotators responsible for detecting signs of depressive symptoms in user posts. Additionally, they must justify their decisions by quoting relevant spans from the original text. This approach aims to replicate the methodology discussed earlier, which outlined the process of finetuning text-to-text models for explanation. 

We explored this strategy in two different ways: first, within a few-shot approach with general purpose models such as \textbf{Vicuna-13B}~{\cite{vicuna2023}} \textbf{GPT-3.5 Turbo}~{\cite{chatgpt}} and \textbf{Mistral-7B-Instruct}~{\cite{jiang2023mistral}}. Second, with domain-specific models that were previously finetuned such as \textbf{MentalLLaMA-chat-13B}~{\cite{yang2023mentalllama}} and \textbf{MentalMistral}. Both finetuned models can follow instructions to generate explanations for the predictions, but they differ in important ways. First, MentalMistral is a QLoRA~{\cite{dettmers2023qlora}}-finetuned version of the Mistral-7B-Instruct model that we implemented with the same symptom-based data as the rest of our models, whereas MentalLLaMA is based on LLaMA2-chat-13B~{\cite{touvron2023llama}}. Their strategies to identify depression also vary: MentalLLaMA focuses on assessing depression by classifying users into depressed and non-depressed categories, while MentalMistral uses clinical indicators to assess depression at the symptom level. Furthermore, the nature of their generated explanations are also different. While MentalLLaMA generates free-form explanations, all our methods, including MentalMistral, are specifically tuned to generate explanations from fragments of the original text.

\section{Experiments}

In this section, we describe the experiments conducted to evaluate the effectiveness of our approaches for symptom detection and explanation over social media content. We present the datasets utilized, followed by a description of the experimental configurations and training details. Subsequently, we explore the metrics employed to measure the performance of our methods, encompassing both offline evaluations and those involving human experts.

\subsection{Datasets}
\label{subsec:explainability_dataset}

In this work, we considered the symptoms of the BDI-II clinical questionnaire~\cite{Beck1996}. The BDI-II covers 21 recognized symptoms, such as pessimism, sleep problems or self-dislike. We have decided to align with the BDI-II symptoms by its widespread adoption in clinical practice~\cite{jackson2016beck} and its presence in prior literature concerning depression detection on the Internet~\cite{crestani2022early,PEREZ2022102380,naseem2022early,ijcai2022p725,parapar2023erisk}. We obtained the symptom evidence sentences from the BDI-Sen~\cite{perez2023sigir} and PsySym~\cite{zhang-etal-2022-symptom} resources. Both datasets consist of symptom-annotated sentences for depression sourced from the Reddit platform. BDI-Sen comprises relevant sentences related to the $21$ symptoms covered in the BDI-II, while PsySym includes $14$ main symptoms covered within the Diagnostic and Statistical Manual of Mental Disorders (DSM-5)~\cite{nuckols2013diagnostic}. Table~\ref{tab:datasets_statistics} presents the statistics of the positive posts from the datasets. In addition to positive posts, we also included control instances in the experiments. We selected random control sentences from the datasets, totalling 1998 control posts. In Section~\ref{sec:ethical-implications}, we comment on the anonymization and ethical use of the data.

\begin{table}[ht]
	\centering
	\caption{Statistics of the positive instances of our explainability dataset. Average length of posts/explanations are expressed as tokens.}
	\label{tab:datasets_statistics}
	\begin{tabular}{lcc}
		\toprule
		&\textbf{BDI-Sen}&\textbf{PsySym}\\
		\midrule
		Number of posts & 357 & 752 \\
		Number of explanations & 546 & 764 \\
		\noalign{\vskip 0.6ex}\cdashline{1-3}[1pt/5pt]\noalign{\vskip 1.0ex}
		\multirow{1}{*}{Avg. number of expls. per post}  
		&1.53 & 1.02 \\ 
		\multirow{1}{*}{Avg. length of post (in tokens)} 
		&127 & 515 \\
		\multirow{1}{*}{Avg. length of explanation} 
		&13.76 & 13.44 \\ 
		\bottomrule
	\end{tabular}
\end{table}

\setlength{\tabcolsep}{1.1pt}
\begin{table*}[ht]
	\centering
	\caption{Settings considered in our experiments by combining the BDI-Sen and PsySym datasets.}
	\label{tab:experimental_settings}
	\begin{tabular}{ccccccccc}
		\toprule
		\multirow{2.3}{*}{\textbf{Setting}} && \multicolumn{3}{c}{\textbf{Training}} && \multicolumn{3}{c}{\textbf{Test}} \\
		\cmidrule(lr){3-5} \cmidrule(lr){7-9}
		&& {\small Dataset} & \textbf{$\oplus$/$\ominus$} & {\small Avg. expls.} && {\small Dataset} & \textbf{$\oplus$/$\ominus$} & {\small Avg. expls.}\\
		\midrule
		\textit{B-B} && {\small BDI-Sen} & 285/285 & 1.48 && {\small BDI-Sen} & 72/359 & 1.72\\
		\textit{B-P} && {\small BDI-Sen} & 285/285 & 1.48 && {\small PsySym} & 151/753 & 1.02\\
		\textit{P-P} && {\small PsySym} & 601/601 & 1.01 && {\small PsySym} & 151/753 & 1.02\\
		\textit{P-B} && {\small PsySym} & 601/601 & 1.01 && {\small BDI-Sen} & 72/359 & 1.72 \\
		\textit{M-M} && {\small Mix} & 886/886 & 1.16 && {\small Mix} & 223/1112 & 1.25 \\
		\bottomrule
	\end{tabular}
\end{table*}
\setlength{\tabcolsep}{5pt}

The datasets comprise two different data sources. Thus, for our experiments, we can define five different settings for train and test sets by combining both resources. Initially, for both BDI-Sen and PsySym, we split their data 80-20 into training and test sets. Then, we fill these splits with control samples from PsySym with ratios 1:1 for training set and $\approx$ 1:5 for the test set\footnote{We include more control sentences in the test set to replicate real scenarios, where there are many more negatives examples than positive ones.}. Table~\ref{tab:experimental_settings} illustrates the five different settings we consider, where we test each dataset on itself, and also using as test the other one to evaluate if it can generalize correctly. In the last setting (\textit{M-M}), we merge both datasets.

\subsection{Experimental Settings}

\subsubsection{Single-step.}

For our single-step strategy, we finetuned two distinct pretrained text-to-text models: \textbf{WT5} and \textbf{WBART}. 
For the WT5 model, we followed the training protocol from Narang et al.~\cite{narang2020wt5}, applying 40 epochs of fine-tuning to the {\small \texttt{t5-large}} model. This adaptation was for our \textit{classify + explain} task, utilizing the AdaFactor optimizer with a constant learning rate of $0.001$. The input and output sequence lengths were set to 2048 and 512 tokens, respectively. In the case of WBART, we fine-tuned the {\small \texttt{bart-large}} model, adhering to the default learning rate and hyperparameters. Due to computational limitations, the maximum lengths for the source and target sequences were reduced to half of those used in WT5. 

Additionally, we employed the conversational LLMs previously mentioned (\textbf{Vicuna-13B}, \textbf{GPT-3.5}, \textbf{Mistral-Instruct}, \textbf{MentalLLaMA}, and \textbf{MentalMistral}) within the single-step framework. All our models were trained and evaluated using NVIDIA A100-SXM4 80GB GPUs. For GPT-3.5, we relied on API calls. All conversational LLMs were directed following the in-context learning strategy described in Subsection \ref{subsec:in_context_learning}, using a balanced set of 30 positive and 30 control samples randomly chosen.

\subsubsection{Two-steps.} In this approach, we first used two BERT-based models for text classification: BERT base, and MBERT base, following the existing implementations from the HuggingFace library without any additional hyperparameter tuning. The learning rate was $2e^{-5}$ during $20$ epochs and a batch size of $32$. We also finetuned a T5 on its {\small \texttt{t5-large}} configuration for text classification by generating labels in textual form. For explanations, we used another {\small \texttt{t5-large}} to explain the symptoms presence. The parameters used to finetune the T5 models were the same as those used for WT5, with same sequence lengths. As a result, we constructed three different variants within this approach: \textbf{T5 + T5}, \textbf{BERT + T5}, and \textbf{MBERT + T5}.

\subsection{Evaluation}
\label{subsec:evaluation}

\subsubsection{Classification}

To evaluate the classification performance of our systems, we consider micro-averaged F1 due to the unbalanced nature of the datasets. By using micro-F1, we account for the importance of each instance, giving more weight to the minority class (positive). Additionally, we include the number of true positives (TPs) to gain insight into the actual number of correctly identified positive instances. This information is particularly significant since our systems only generate explanations for the positive samples.

\subsubsection{Explanation}

LLMs can generate text unsupported by the input and produce inaccurate responses, a phenomenon called ``hallucination''~\cite{ji2023survey}. This is a magnified risk considering the sensitive factor of the mental health domain~\cite{yang2023interpretable}. For this reason, we force extractive explanations and instruct the models to reflect on a given text and provide a decision-supporting fragment. We ensure that our systems are less prone to hallucination than in a free-form explanation setup. In the clinical domain, where the validity and truthfulness of the explanations extracted are crucial~\cite{10.1093-jamiaopen-ooz054}, the generated texts' reliability is essential. To ensure the trustworthiness of the generated explanations, we complement classical offline metrics with \textit{expert-in-the-loop} evaluation performed with three domain experts.

\paragraph{Offline}

We used three offline metrics: ROUGE-L-F1~\cite{lin-2004-rouge}, Corpus BLEU~\cite{papineni-etal-2002-bleu}, and Token F1~\cite{deyoung-etal-2020-eraser}, each offering different evaluation perspectives. All these metrics compare the generated explanations with the text reference considered as golden truth. ROUGE emphasizes content overlap, ensuring that the generated output captures essential information from the references. On the other hand, BLEU focuses on fluency and adequacy, assessing the linguistic quality and alignment with references. Since these metrics effectively measure the quality of the generated text by comparing it against reference hypotheses, ROUGE and BLEU can evaluate the quality of our explanations while still allowing some paraphrasing. This flexibility can be beneficial when the explanation text slightly differs from the ground truth yet conveys the same meaning. Additionally, Token F1 measures which input tokens labelled as an explanation in the ground truth are present in the generated one. Thus, Token F1 score is computed through a token-by-token analysis, identifying if the tokens generated are the same as the references.

\paragraph{Expert-in-the-loop}
\label{subsubsec:expert-in-the-loop}

We also performed an online evaluation using three domain experts to assess the practical utility and clinical relevance of the generated explanations. The evaluators consisted of two psychologists and a speech therapist. To ensure the consistency of judgments, we organized training sessions with the evaluators during which we discussed the criteria for relevance and established a consensus on samples of both positive and negative explanations. An explanation was considered relevant if it successfully provided evidence for the presence of a depressive symptom in the post. For a detailed view of the instructions provided to the assesor please see \ref{appendix:assessment}. We provided the experts with explanations generated by our WT5 model variant under the \textit{M-M} setting (see Table \ref{tab:results}), leading to a total of $209$ explanations being evaluated. We chose this setting because it includes both datasets and selected WT5 since it was the best performing model in this scenario.

\begin{table*}
	\centering
	\caption{Results of our methods in the different proposed settings. F1, TPs, ROUGE, BLEU and TF1 denote micro-F1, number of true positives, ROUGE-L-F1, Corpus BLEU and Token F1. More details about these metrics can be found in Subsection \ref{subsec:evaluation}. All metrics are in the range 0-1, except for TPs, which are an integer. Higher values indicate better performance.}
	\label{tab:results}
	\resizebox{0.85\linewidth}{!}{
		\begin{tabular}{clrrrrrrrrrrr}
			\toprule 
			&  & \multicolumn{5}{c}{\rotatebox{0}{\textbf{Finetuned PLMs}}} & \multicolumn{5}{c}{\rotatebox{0}{\textbf{In-context convers. LLMs}}} \\
			\cmidrule(lr){3-7} \cmidrule(lr){8-12}
			&  & \multicolumn{2}{c}{\rotatebox{0}{Single-step}} & \multicolumn{3}{c}{\rotatebox{0}{Two-steps}} & \multicolumn{3}{c}{\rotatebox{0}{Few-shot}} & \multicolumn{2}{c}{\rotatebox{0}{Finetuned}}\\
			\cmidrule(lr){3-4} \cmidrule(lr){5-7} \cmidrule(lr){8-10} \cmidrule(lr){11-12}
			\textbf{Setting} & \textbf{Metric} & \rotatebox{90}{WT5} & \rotatebox{90}{WBART} & \rotatebox{90}{T5 + T5} & \rotatebox{90}{BERT + T5} & \rotatebox{90}{MBERT + T5} & \rotatebox{90}{Vicuna-13B} & \rotatebox{90}{GPT-3.5} & \rotatebox{90}{Mistral-Instruct} & \rotatebox{90}{MentalLLaMA} & \rotatebox{90}{MentalMistral} \\
			\midrule
			
			\multirow{5}{*}{\textit{B-B}} 
			& F1   & 0.91 & 0.88 & 0.93 & \textbf{0.95} & 0.92 & 0.73 & 0.82 & 0.64 & 0.88 & 0.89 \\
			& TPs  & \textbf{68} & \textbf{68} & 67 & 58 & 65 & 41 & 62 & 68 & 51 & 51 \\
			& ROUGE& 0.62 & 0.60 & 0.69 & \textbf{0.75} & 0.72 & 0.42 & 0.55 & 0.39 & 0.39 & 0.44 \\
			& BLEU & 0.53 & 0.53 & 0.55 & \textbf{0.61} & 0.58 & 0.22 & 0.42 & 0.18 & 0.19 & 0.26 \\
			& TF1  & 0.48 & 0.48 & 0.54 & \textbf{0.60} & 0.58 & 0.34 & 0.45 & 0.36 & 0.37 & 0.31 \\
			\midrule
			
			\multirow{5}{*}{\textit{B-P}} 
			& F1   & 0.87 & 0.86 & 0.86 & 0.91 & 0.92 & 0.78 & 0.87 & 0.66 & 0.92 & \textbf{0.93} \\
			& TPs  & 125 & 134 & 79 & 79 & 144 & 140 & \textbf{151} & 150 & 131 & 141 \\
			& ROUGE & 0.31 & 0.22 & \textbf{0.35} & 0.34 & 0.29 & 0.14 & 0.17 & 0.12 & 0.12 & 0.17 \\
			& BLEU  & 0.22 & 0.15 & \textbf{0.23} & 0.22 & 0.18 & 0.06 & 0.08 & 0.07 & 0.07 & 0.08 \\
			& TF1   & 0.10 & 0.10 & \textbf{0.12} & \textbf{0.12} & 0.10 & 0.11 & 0.10 & 0.12 & 0.17 & 0.09 \\
			\midrule
			
			\multirow{5}{*}{\textit{P-P}} 
			& F1   & \textbf{0.98} & \textbf{0.98} & \textbf{0.98} & 0.97 & \textbf{0.98} & 0.69 & 0.94 & 0.80 & 0.93 & 0.96 \\
			& TPs  & 143 & 149 & 145 & 150 & 149 & 150 & \textbf{151} & 150 & 147 & 144 \\
			& ROUGE & 0.53 & \textbf{0.56} & 0.45 & 0.47 & 0.45 & 0.15 & 0.21 & 0.16 & 0.16 & 0.17 \\
			& BLEU  & 0.43 & \textbf{0.53} & 0.34 & 0.36 & 0.35 & 0.08 & 0.11 & 0.12 & 0.08 & 0.08 \\
			& TF1   & 0.47 & \textbf{0.51} & 0.38 & 0.38 & 0.38 & 0.09 & 0.11 & 0.08 & 0.09 & 0.08 \\
			\midrule
			
			\multirow{5}{*}{\textit{P-B}} 
			& F1   & 0.89 & 0.89 & 0.89 & \textbf{0.91} & 0.90 & 0.62 & 0.85 & 0.73 & 0.85 & 0.84 \\
			& TPs  & 35 & 35 & 40 & 48 & 42 & \textbf{53} & 49 & 47 & 51 & 29 \\
			& ROUGE & 0.61 & \textbf{0.71} & 0.61 & 0.57 & 0.61 & 0.36 & 0.53 & 0.48 & 0.47 & 0.48 \\
			& BLEU  & 0.61 & \textbf{0.68} & 0.60 & 0.57 & 0.58 & 0.19 & 0.34 & 0.34 & 0.33 & 0.31 \\
			& TF1   & 0.46 & \textbf{0.51} & 0.45 & 0.42 & 0.45 & 0.23 & 0.36 & 0.32 & 0.35 & 0.30 \\
			\midrule
			
			\multirow{5}{*}{\textit{M-M}} 
			& F1   & 0.95 & 0.94 & 0.95 & 0.95 & \textbf{0.96} & 0.59 & 0.91 & 0.45 & 0.77 & 0.88 \\
			& TPs  & 209 & 211 & 206 & 211 & 211 & 195 & 200 & \textbf{215} & 191 & 191 \\
			& ROUGE & \textbf{0.57} & 0.53 & 0.54 & 0.54 & 0.54 & 0.22 & 0.28 & 0.25 & 0.23 & 0.24 \\
			& BLEU  & \textbf{0.54} & 0.47 & 0.51 & 0.51 & 0.51 & 0.10 & 0.17 & 0.15 & 0.09 & 0.19 \\
			& TF1   & \textbf{0.50} & 0.48 & 0.46 & 0.46 & 0.45 & 0.14 & 0.19 & 0.17 & 0.17 & 0.14 \\
			
			\bottomrule
		\end{tabular}
	}
\end{table*}

\begin{table}[ht]
	\centering
	\caption{Metric averages per method and proposed setting. F1, TPs, ROUGE, BLEU and TF1 denote micro-F1, number of true positives, ROUGE-L-F1, Corpus BLEU and Token F1. More details about these metrics can be found in Subsection \ref{subsec:evaluation}. All metrics are in the range 0-1, except for TPs, which are an integer. Higher values indicate better performance.}
	\label{tab:avgs_combined}
	
	\begin{subtable}{\columnwidth}
		\centering
		
		\begin{tabular}{rcccc}
			\toprule
			Method & \textbf{F1} & \textbf{ROUGE} & \textbf{BLEU} & \textbf{TF1} \\
			\midrule
			WT5                 & 0.91 & \textbf{0.53} & \textbf{0.47} & 0.40 \\
			WBART               & 0.91 & 0.52 & \textbf{0.47} & \textbf{0.42} \\
			\noalign{\vskip 1ex}
			T5 + T5             & 0.92 & \textbf{0.53} & 0.45 & 0.39 \\
			BERT + T5           & \textbf{0.94} & \textbf{0.53} & 0.45 & 0.40 \\
			MBERT + T5          & \textbf{0.94} & 0.52 & 0.44 & 0.39 \\
			\noalign{\vskip 1ex}
			Vicuna-13B          & 0.68 & 0.26 & 0.13 & 0.18 \\
			GPT-3.5             & 0.88 & 0.25 & 0.22 & 0.24 \\
			Mistral-Instruct    & 0.66 & 0.28 & 0.17 & 0.21 \\
			\noalign{\vskip 1ex}
			MentalLLaMA         & 0.87 & 0.24 & 0.15 & 0.23 \\
			MentalMistral       & 0.90 & 0.30 & 0.18 & 0.18 \\
			\bottomrule
		\end{tabular}
		\subcaption{Per method.}
		\label{subtab:avgs_per_methods}
	\end{subtable}
	
	\medskip
	
	\begin{subtable}{\columnwidth}
		\centering
		
		\begin{tabular}{rrrrrr}
			\toprule
			Setting & \textbf{F1} & \textbf{TPs} & \textbf{ROUGE} & \textbf{BLEU} & \textbf{TF1} \\
			\midrule
			\textit{B-B}    & 0.85 & 60 & \textbf{0.56} & 0.41 & \textbf{0.45} \\
			\textit{B-P}    & 0.86 & 127 & 0.22 & 0.14 & 0.11 \\
			\textit{P-P}    & \textbf{0.92} & 148 & 0.33 & 0.25 & 0.26 \\
			\textit{P-B}    & 0.84 & 43 & 0.54 & \textbf{0.46} & 0.39 \\
			\textit{M-M}    & 0.84 & 204 & 0.38 & 0.32 & 0.32 \\
			\bottomrule
		\end{tabular}
		
		\subcaption{Per setting.}
		\label{subtab:avgs_per_setting}
	\end{subtable}
\end{table}

\section{Results and Discussion}

In this section, we present the results of the experiments designed to address the four research questions posed in the introduction.

\subsection{\cref{rq1}: Can we train transformers models to detect and explain depressive symptoms?}

\subsubsection{Symptoms classification.} The results presented in Table~\ref{tab:results} confirm that it is possible to train transformer architectures to detect traces of depression symptoms in social media posts. Across all the different settings, we observe that the worst performing finetuned PLM achieves a classification accuracy of $0.86$, which is the case for the WBART and T5 + T5 methods for the \textit{B-P} setting. On the other hand, in-context conversational LLMs have more difficulties in detecting the presence of depression symptoms within a few-shot approach, and in some cases their performance drops to an accuracy of $0.45$, which is the case of the Mistral-Instruct method for \textit{M-M} setting. However, conversational LLMs show the ability to learn this task easily when finetuned. Looking at the MentalMistral method, which is a finetuned version of Mistal-Instruct at the symptom level, we can see that it improves its classification accuracy from $0.45$ to $0.88$ and from $0.66$ to $0.93$ for the \textit{M-M} and \textit{B-P} settings, respectively. It is particularly noteworthy that for the latter setting, MentalMistral is the method that best detects the symptoms of depression. Comparatively, MentalLLaMA, another fine-tuned LLM designed for broader mental health conditions beyond depression, shows slightly inferior performance compared to MentalMistral.

To better understand how well the different methods detect depression symptoms across settings, Table~\ref{subtab:avgs_per_methods} contains a column with the F1 average across every setting. The values are consistent with the analysis just presented. On the one hand, all finetuned PLMs have a similar performance in terms of detection, which is slightly higher in the case of the two-step pipeline. It is also clear that the in-context conversational LLMs are a step behind, with the Mental-Instruct method being the worst performer (average F1 of $0.66$), but that they improve significantly when finetuned, either at the symptom level (MentalMistral) or not (MentalLLaMA).

We also show in Table~\ref{subtab:avgs_per_setting} a comparison of the performance of our methods across all the dataset settings. We observe the best classification figures in the \textit{P-P} setting (training and testing on the PsySym dataset), with all finetuned PLMs achieving F1 values above $0.97$ and all methods averaging an F1 value of $0.92$. This numbers also provide insights into how the models generalise when tested on a different dataset (the \textit{B-P} and \textit{P-B} settings). We observe bad generalisation in the case of the \textit{P-B} setting. For instance, looking at specific models in Table~\ref{tab:results} the T5 + MBERT model achieved an F1 of 0.98 when trained and tested on the PsySym dataset (\textit{P-P} setting). Meanwhile, its performance dropped to $0.90$ when tested on the BDI-Sen dataset (\textit{P-B} setting). However, in the \textit{M-M} setting, the trained models obtained an F1 greater than $0.94$.

\begin{figure}[ht]
	\centering
	\includegraphics[width=1.0\linewidth]{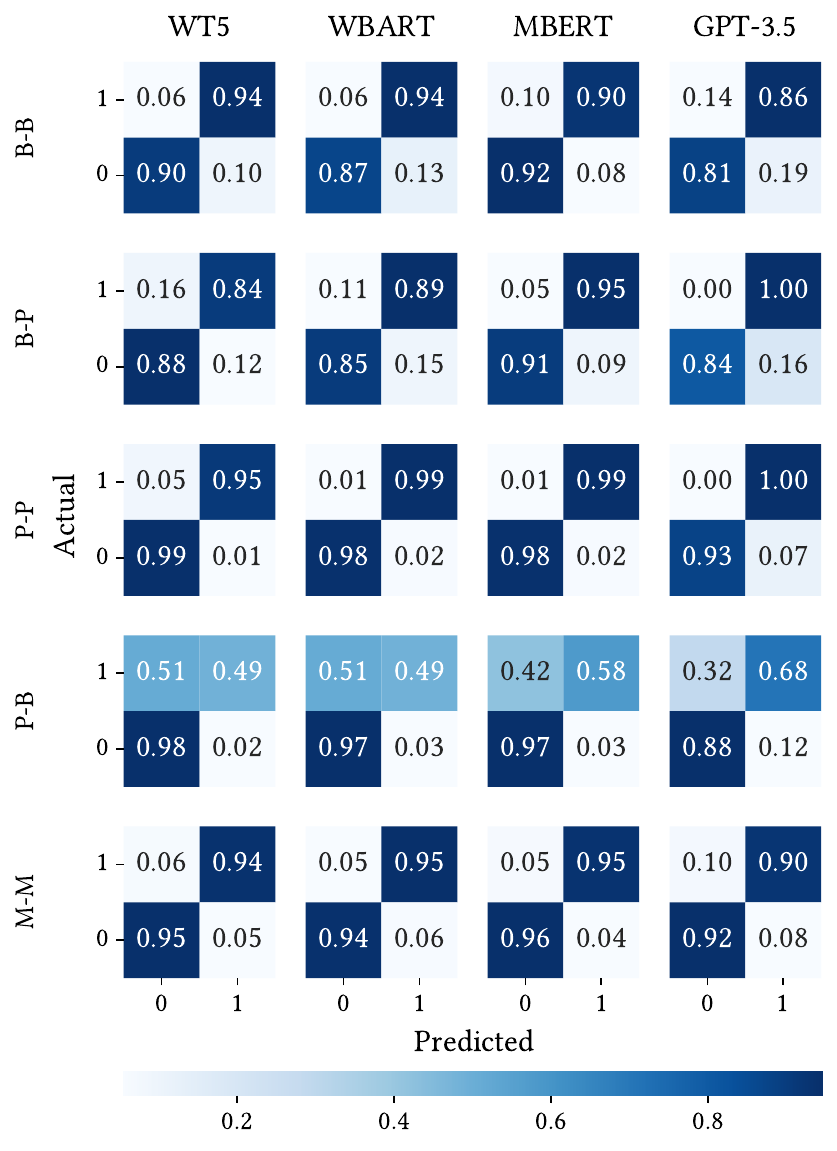}
	\caption{Confusion matrices showing the predictions accuracy for the proposed settings and WT5, WBART, MBERT and GPT-3.5 systems.}
	\label{fig:confusion_matrices}
\end{figure}

To further analyse this generalisation problem, we illustrate the distribution of true and false predictions of the systems in Figure~\ref{fig:confusion_matrices}. 

As observed in the confusion matrices\footnote{Due to clarity and space constraints, not all systems are shown.}, the models have a high ratio of true positives, with the majority identifying over $90\%$ of the positive instances. Furthermore, the ratio of false negatives is remarkably small. However, we can again observe significant deviation in these trends in the matrix for the \textit{P-B} setting. Here we can see the reason for that problem: the PsySym dataset covers 14 symptoms of depression, while BDI-Sen includes all 21 symptoms of the BDI-II. In the \textit{P-B} setting, not all symptoms of BDI-Sen have been seen during training, leading to a substantially higher false positive number. This finding highlights the potential generalisation issues these models may encounter regarding the data used for training. False negatives are a crucial metric for clinical applications. In clinical practice, false negatives carry great significance. Failing to identify an individual at risk has far more severe implications than falsely identifying a healthy person (false positives).

{
	\begin{table*}
		\centering
		
		\caption{Each row presents an example of positive predictions in the \textit{M-M} setting using the WT5 model. Explanations are \tcbox{highlighted} as judged by the assessors. The \textit{Relevant} column displays the relevance labels after the human assessment. Samples have been paraphrased for privacy and shortened for clarity.}
		\label{tab:explanations-selected}
		\begin{tabular}{p{0.83\linewidth} c}
			\toprule
			\multicolumn{1}{c}{Input post} & Relevant \\
			\midrule
			
			{\small \texttt{(...)} I want to start a business one moment, then pay out my IRA and travel throughout Europe. I do not comprehend who I am. \tcbox{\textbf{My short-term memory is terrible, and I can not concentrate}}. I'm unsure of what to do.  You guys are going to advise some really fantastic actions for me to pursue, but ultimately I lack the willpower or energy to carry out your advice.}  & 
			\multirow{3}{*}{1} \\
			\midrule
			
			{\small Recently, I have been having a lot of difficulty with this. \tcbox{\textbf{I have been depressed, worried, or ill since I was a child}}. Like my youth has been taken from me. As a last-ditch effort to feel like a person once more, I am actually considering seeing a naturopathic physician. Anyway, I hope you soon feel better.} & \multirow{3}{*}{1}\\
			\midrule
			
			{\small I abhor myself to the core. Even reading back through Reddit postings I posted a few days ago makes me want to commit suicide. \tcbox{\textbf{I am such a disgusting waste of life — useless, unproductive,}} \tcbox{\textbf{and with a future that is already in uncertainty}}. And the fact that Im feeling this way on spring break is something I detest so fiercely. \texttt{(...)}} & \multirow{3}{*}{1}\\
			\midrule
			
			{\small \texttt{(...)} how EMTs and first responders are looking after them, and how those individuals should persevere to witness another day. \tcbox{\textbf{Nonetheless, I find myself unable to avoid fall into thoughts}} \tcbox{\textbf{about someone's death}}. This triggers memories of my own experiences and how I might find myself in that person's position \texttt{(...)}} & \multirow{3}{*}{0}\\
			
			\bottomrule
		\end{tabular}
	\end{table*}
}

\subsubsection{Symptoms explanation.} 
\label{subsec:symptoms-explanation}

Regarding explanation quality, offline metrics and human evaluation results indicate promising results. Once again, finetuned PLMs outperform the most recent conversational instructed LLMs, which can be seen in columns ROUGE, BLEU and TF1 of Table~\ref{subtab:avgs_per_methods}. It is interesting to note that, in contrast to symptom detection, modern LLMs do not significantly improve the quality of their explanations after finetuning. This idea is consistent with the results presented for \ref{rq2}, where it is shown that the quantity and quality of examples is much more critical for explanation than for classification supporting the idea that generating explanations is more challenging than the classification task. As previously stated ($\S$\ref{subsec:notes-explanations}), it is crucial to bear in mind that we generate explanations only for positive cases. Hence, the explanation quality numbers have to be jointly considered with the number of true positives (refer to the TPs rows in Table~\ref{tab:results}). When considering the offline metrics (ROUGE, BLEU, and TF1), the fairest comparison can be made in the \textit{M-M} setting, since it is the largest setting and provides a similar number of TPs cases for all models. Here, WT5 emerges as the top performer, achieving a ROUGE-L F1-score of $0.57$. The two-step models also perform closely to the single-step models (WT5 and WBART) in nearly all settings. In the \textit{B-B} setting, where the numbers of true positives are also roughly similar, they outperform the single-step models.

In terms of human evaluation, we presented the explanations generated by our WT5 model in the \textit{M-M} setting to three domain experts, with $209$ explanations provided to each of them. Table~\ref{tab:explanations-selected} provides four examples of positive predictions along with their generated explanations. The first three rows correspond to relevant explanations, since the three human assessors considered them as relevant. The last row shows a non-relevant explanation. In this case, while it appears to be topically related to a symptom of depression (\textit{Suicidal ideas}), the experts classified it as non-relevant, as it does not provide any information about the symptom. Overall, the three assessors found $73$\%, $53$\%, and $77$\% of the explanations relevant and clinically useful, resulting in an average of 68\%. A more granular inspection of the assessors' annotations indicates that, out of $209$ explanations, $154$ were deemed clinically useful by at least two of the three assessors. In terms of total consensus, which is an important factor as it unambiguously confirms or rejects the practical utility of an explanation, we find that assessors agreed on the non-relevance of $25$ explanations and the relevance of $86$. Regarding the inter-annotator agreement among domain experts in our study, we computed a pairwise agreement to measure the percentage of cases where the three assessors reached a consensus in their annotations. The observed agreement values among the pairs were $76$\%, $66$\%, and $65$\%, yielding an average agreement of $69$\% regarding the clinical relevance of the explanations. This level of concordance aligns with similar studies in the field~\cite{perez2023sigir,zhang-etal-2022-symptom,macavaney-etal-2018-rsdd}, considering the challenges associated with annotating in the complex domain of mental health~\cite{mowery2017understanding}.

\subsection{\cref{rq2}: How many labels are needed to generate good explanations?}

\begin{figure*}[ht]
	\centering
	\includegraphics[width=\linewidth]{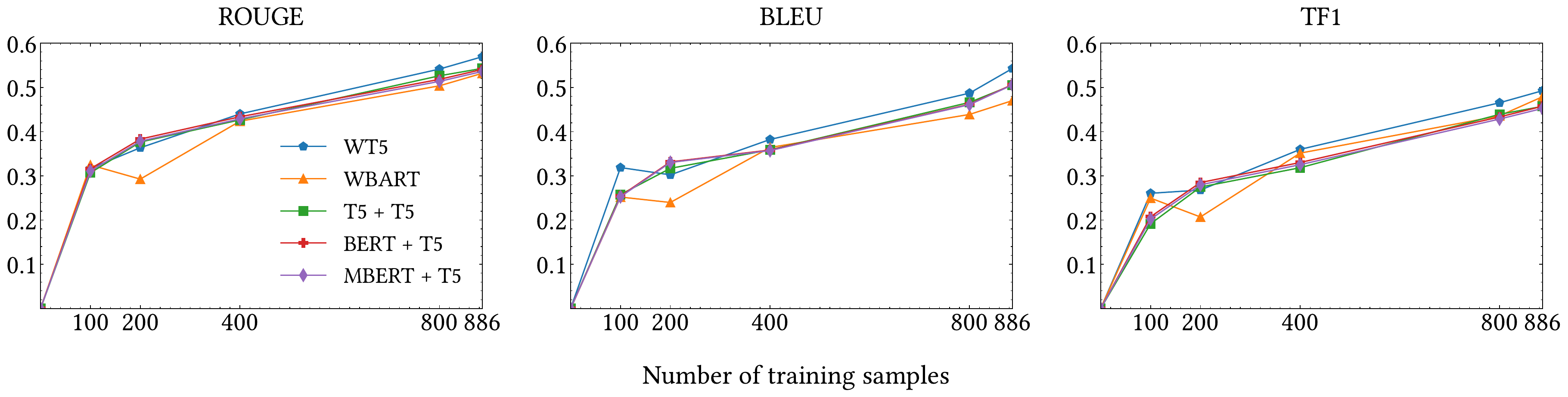}
	\caption{Offline metrics in relation to the quantity of samples kept in the training data for the M-M setting}
	\label{fig:limited_explanations}
\end{figure*}

Since collecting data labelled by domain experts in the mental health domain on social media is a costly process~\cite{harrigian2020models,macavaney-etal-2018-rsdd}, we investigate here how many training samples are needed to generate quality explanations. To analyze this aspect, we designed two different experimental setups: First, in~\ref{subsubsec:sample_quantity_impact}, we investigate the effect of the number of labelled examples on the quality of explanations for the experimental setup used in this paper. Then, in~\ref{subsubsec:external_dataset_validation}, we validate the results of this experiment adapting and including an additional external dataset.

\subsubsection{Impact when training with limited explanations}
\label{subsubsec:sample_quantity_impact}

To assess how the quantity of training samples would affect our experiments, we trained the PLMs with a gradually increasing number of examples using our larger setting, denoted as \textit{M-M}. This setting includes $886$ explanations in the training set, and we compared our models' performance using subsets of $100$, $200$, $400$, and $800$ training samples. Figure~\ref{fig:limited_explanations} presents the results across the three evaluation metrics (ROUGE, BLEU and TF1) that measure the quality of explanations generated. As shown in the Figure, there is a nearly linear improvement with respect to the number of training instances for all metrics and models. For instance, when our models are trained with only 100 explanations, we achieve approximately half the performance compared to the entire training split. For completeness, we also investigated here the performance of the classification task, where we observed a more robust behaviour. For instance, using only 200 samples, all models consistently achieved a good F1 score, being higher than $0.9$. In line with previous literature~\cite{narang2020wt5}, we find that the number of training samples affects the quality of the explanations more than the classification results.

\subsubsection{Assessing explanation quality stabilization when varying number of training samples}

\label{subsubsec:external_dataset_validation}

\begin{figure*}[ht]
	\centering
	\includegraphics[width=\linewidth]{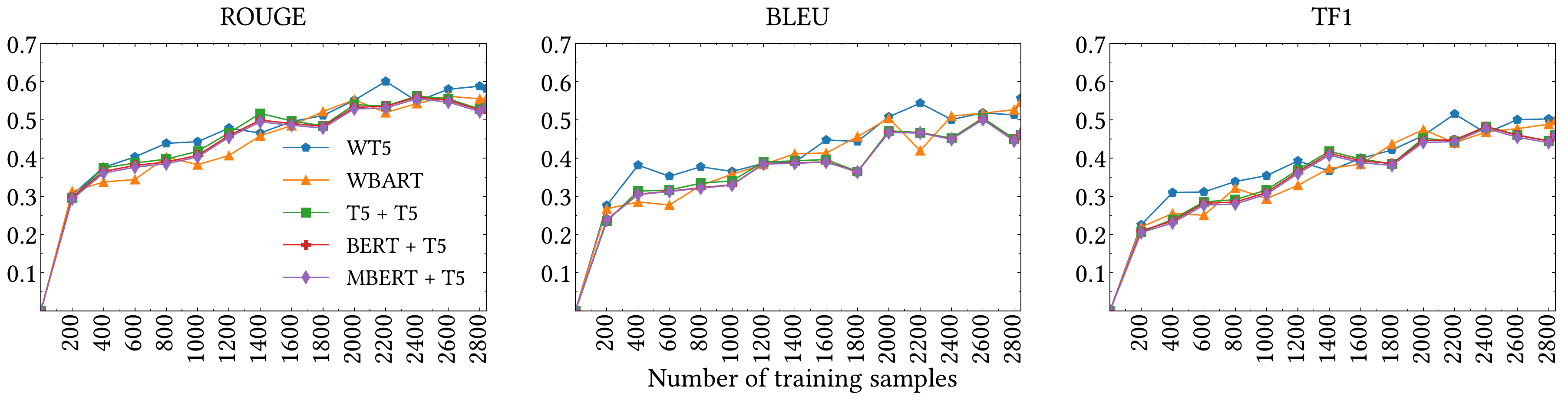}
	\caption{External dataset validation incorporating DepreSym training samples.}
	\label{fig:limited_explanations_depresym}
\end{figure*}

Following the above experimentation described in~\ref{subsubsec:sample_quantity_impact}, we further investigated this aspect by conducting an additional experiment adapting the DepreSym~\cite{perez2023depresym} resource, an external dataset sourced from Reddit. This dataset also contains sentences that indicate the presence of markers for depression symptoms, allowing us to collect the user posts and then use these sentences as explanations. Using this resource, we shuffled these new explanations within the previous training set of the \textit{M-M} setting and evaluated the explanation metrics every $200$ training examples for the original \textit{M-M} test set. Thus, we added $1956$ new training examples, but kept the previous test set. Figure~\ref{fig:limited_explanations_depresym} shows our results. Looking at the performance of our models, we can see that the values of the metrics continue to improve after $800$ training instances. However, around $2000$ instances, we found a trend towards stabilisation of performance. Therefore, we can observe that after a certain point incorporate additional training samples does not guarantee an improvement on the performance. 

\subsection{\cref{rq3}: Single-step vs Two-steps}

Through our experimentation, we conducted a comparative analysis of finetuned PLMs to perform the \textit{classify + explain} tasks in a single step (WT5 and WBART) against those operating in two separate steps (T5 + T5, BERT + T5, MBERT + T5). Regarding the classification task, the two-step models exhibited a slight advantage, as reflected in column F1 from Table~\ref{subtab:avgs_per_methods}. Specifically, the BERT-based models consistently outperformed the other models, with BERT and MBERT achieving the best classification performance in four of the five settings. Regarding explanation, to ensure a fair comparison, we focused on settings with a similar number of true positives (TPs), which led us to consider the \textit{P-P} and \textit{M-M} settings. In both cases, the models with the best explanations were WT5 and WBART.

In the context of our investigation, we can see that both the single-step and two-step approaches show promising potential for the classification and explanation tasks. However, we can see slight differences of the models in the two tasks. Looking at the results, we observe that, single-step systems often yield superior results in terms of explanation, even if their classification performance is marginally lower. This aligns with prior research suggesting that consolidating classification and explanation into a single step can enhance the quality of explanations~\cite{feder-etal-2022-causal, BARREDOARRIETA202082, narang2020wt5}. Under the single-step paradigm, classification and explanation are treated as an interconnected whole rather than discrete operations. This integration may lead to more cohesive and informative explanations, underscoring the importance of considering the aggregated nature of the task. Regarding the slight better performance of the two-steps models in terms of classification, this also aligns with prior work, since BERT-based models are still predominant in classification for similar scenarios~\cite{yang2023interpretable}. 

\subsection{\cref{rq4}: In-context Conversational instructed LLMs performance}

When evaluating the results achieved by conversational LLMs, it is important to distinguish between general purpose models used within a few-shot approach (Vicuna-13B, GPT-3.5 and Mistral-Instruct) and domain-specific finetuned versions such as MentalLLaMA and MentalMistral. The overall results presented in Table~{\ref{tab:results}} show that all the instructed LLMs achieve lower performance than our finetuned PLMs, especially for the explanation task. Table~{\ref{subtab:avgs_per_methods}} reveals that it is GPT-3.5 method that comes closer to the best-performing systems among the general purpose LLMs in terms of classification but, remains far behind in terms of explanations quality. Moreover, our experiments also show that conversational LLMs benefit greatly from finetuning, whether they were finetuned at the symptom level (MentalMistral) or not (MentalLLaMA). This leap in performance can be demonstrated by comparing the results of MentalMistral with those of its base model, Mistral-Instruct. MentalMistral experiences a notorious boost and establishes itself as the model with the best classification results among the conversational LLMs. However, as noted in §\ref{subsec:symptoms-explanation}, even these fine-tuned LLMs face challenges in generating good explanations, and can provide inappropriate and inaccurate explanations due to the hallucination effect~\cite{ji2023survey}. In order to study this phenomenon in the context of our experiments, we calculated the percentage of cases where the generated explanations do not exactly match the original text. For the GPT-3.5 model and the \textit{M}-\textit{M} reference setting, this occurs in only $1.5$\% of cases. This low rate suggest a high degree of precision in the model's output, exemplifying its effectiveness in generating extractive explanations.

Despite these results, it is important to consider that we only used 30 positive and negative examples to guide these systems through in-context learning~\footnote{Models have a maximum sequence length which implies a limit on the number of samples.}. The limited exposure to examples for this new task might have contributed to their relatively lower performance. Moreover, recent studies have shown the variability in the performance of these models depending on the prompts quality~\cite{zhao2021calibrate,7580601}. The prompt we defined for all of our experiments is listed on~\ref{appendix:prompt}, and we plan to explore the impact of different prompt strategies in these particular scenarios.


\section{Conclusions and future work}

In this paper, we presented the use of different text-to-text pipelines to classify and explain the presence of depressive symptoms on social media. Our models, in addition to detecting relevant social media posts related to depressive symptoms, also explain in natural language their decisions. Leveraging two datasets containing sentences indicative of depressive symptoms, we evaluated our model variants using a wide variety of offline metrics and expert-in-the-loop evaluations. Our results are promising in both classification and explanation tasks. Within our methods, we compared single-step approaches (where classification and explanation are performed by the same model) against two-step approaches (where the two tasks are performed by separate models). Moreover, our analysis extended to examining the capabilities of recent conversational LLMs in detecting and explaining depressive symptoms. Our study enriches the expanding field of explainable AI in mental health applications. By offering healthcare systems models capable of explaining their decision-making process, we allow clinicians to understand the automated reasoning and build trust in the outputs of these models.

While we presented an extractive approach for generating explanations, our discussions with domain experts have revealed the potential to create abstract explanations, matching the way one expert would explain the existence of depressive symptoms to another. Moreover, it might be advantageous to identify the precise symptom present in positive posts, leaving behind the actual binary paradigm and defining a new multi-class approach. An important direction for future work involves incorporating temporal information into our models. We are planning to integrate the publication dates of Reddit posts, enabling our systems to process data in temporal batches. This approach would facilitate weekly or monthly analysis, thereby improving our decision-making processes. Such temporal integration is valuable for providing clinicians with insights and evidence of the evolution of depressive symptoms in individuals over time. Additionally, this incorporation would allow us to conduct population-level analyses, examining symptom prevalence during specific periods, such as the COVID-19 pandemic~\cite{first2021covid}. This temporal perspective may help to understand and respond to fluctuating mental health trends in the digital era. As a limitation, we are also aware that, although we use two different datasets, the data obtained comes from a single social media platform. In future research, extending our approach across diverse platforms, languages, and cultural contexts would be valuable to evaluate its cross-cultural applicability. Overall, we believe that, the creation of models capable of explaining the detection of clinical markers, is an important first step towards developing sophisticated natural language explanations for depression detection and related analysis on social media.

\section{Ethical and Environmental Aspects}
\label{sec:ethical-implications}

The data used in our study were obtained from publicly accessible sources, adhering to the exempt status under title 45 CFR §46.104. The use of BDI-Sen, PsySym and DepreSym datasets was accomplished in full compliance with their respective data usage policies. To maintain privacy, we implemented measures to ensure that any personal information was unidentifiable and all users remained anonymous. The data used were sourced from Reddit, and we strictly conformed to all terms specified by this platform. For the human evaluation, even though the assessors were domain experts, they were not subjected to any imposed time restrictions, and reported no adverse effects post-evaluation. Importantly, it must be emphasized that the systems described in this study are intended to complement the work of healthcare professionals, not replace them. The development of such technologies necessitates a cautious approach, with a continuous emphasis on their ethical use and a firm respect for patient privacy and autonomy.

The experiments for this study were conducted using our private infrastructure, with a carbon efficiency of 0.432 kgCO$_2$eq/kWh, which reflects the OECD's 2014 yearly average. The resources utilised included 10 hours of computation on an RTX A6000 device (with a TDP of 300W) and 50 hours on an A100 PCIe 40/80GB device (with a TDP of 250W). Total emissions are estimated to be $8.43$ kgCO$_2$eq. To provide some perspective, this is equivalent to driving an average car for $34$ kilometres. These figures were determined with the assistance of the MachineLearning Impact Calculator~\cite{lacoste2019quantifying}.

\section{Acknowledgments}

This work has received support from projects: PLEC2021-007662 (MCIN/AEI/10. 13039/501100011033 Ministerio de Ciencia e Innovación, European Union NextGenerationEU/PRTR) and PID2022-137061OB-C21 (MCIN/AEI/10.13039/501100011033/, Ministerio de Ciencia e Innovación, ERDF A way of making Europe, by the European Union); Consellería de Educación, Universidade e Formación Profesional, Spain (grant number ED481A-2024-079 and accreditations 2019–2022 ED431G/01 and GPC ED431B 2022/33) and the European Regional Development Fund, which acknowledges the CITIC Research Center.

We would also like to thank Desireé Pombar, Silvia López-Larrosa and Laura Hermo for their efforts in evaluating the results. Their work allowed us to assess the practical utility and clinical relevance of the explanations generated.

\begin{appendices}

{
	\onecolumn
	\section{Assessment guidelines}
	\label{appendix:assessment}
	
	\setlength{\parindent}{0pt}
	
	Considering different depressive symptoms based on the BDI questionnaire (e.g: Sleep problems, Suicidal Ideation, Sadness, Pessimism…), the original task is that, from a Reddit publication, the predicted explanations must give relevant information to some kind of depressive symptomatology.\\
	
	\textbf{Important Notes}\\
	
	A relevant explanation should provide some information about the state of the individual related to a depressive symptom. It is not necessary that the exact same words are used for the explanation, but they must correspond to what was described in the original post. Each post can have more than one explanation. In this case, if one of them is relevant, this is enough to assess it as relevant.\\
	
	\textbf{Examples (some original posts and the correct explanation)} \\
	
	\begin{itemize}
		
		\item Relevant explanation\\
		\begin{itemize}
			\item Post: \say{I'm very concerned about my future. I'm in uni atm, and I'm terrified of not getting a good job, not being able to have kids, not passing my exams, not having enough money to buy my own house, take care of my kids (I'm staying hopeful) or do the things I want to do. I'm basically terrified of not being happy with my life}.
			\item Explanation: \say{I'm basically terrified of not being happy with my life}.\\
		\end{itemize}
		
		\item Relevant explanation\\
		\begin{itemize}
			\item Post: \say{My problem was that i tried to do everything at the same time. Its important to take a breath few times a day. Like for me I'm a student and after school my plan was to, practice programming(python), doing my website project, reading a book about programming and practicing art. I burnt out and I am still unmotivated and feeling like shit. Don't do too many things at the same time}.
			\item Explanation: \say{I burnt out and i still am unmotivated and feeling like shit}.\\
		\end{itemize}
		
		\item Relevant explanation\\
		\begin{itemize}
			\item Post: \say{Yeah. I feel like I don't even get excited about things anymore}.
			\item Explanation: \say{I feel like I don't even get excited about things anymore}.\\
		\end{itemize}
		
		\item Non-Relevant explanation\\
		\begin{itemize}
			\item Post: \say{Even if it is, big fuckin' woop. I liked the movie, and this event was fun. Who cares who organized it?}.
			\item Explanation: \say{I liked the movie, and this event was fun.}.\\
		\end{itemize}
		
		\item Non-Relevant explanation\\
		\begin{itemize}
			\item Post: \say{Nope. We are from Germany}.
			\item Explanation: \say{We are from Germany}.\\
		\end{itemize}
	\end{itemize}

	Your job is to assess if the explanations given to a specific post are informative about some kind of depressive symptomatology. The relevance grades are:\\
	
	\begin{itemize}
		\item Relevant (1): A relevant explanation should be relevant to a depressive symptom, and it must correspond to the content written in the original publication.\\
		\item Non-relevant (0): A non-relevant sentence does not address any topic related to a depressive symptom, or the explanation is not informative about the original publication.\\
	\end{itemize}
	
	The Excel sheet contains three columns (Post, Explanation and Relevant Explanation). The column \say{Post} contains the original post, the \say{Explanation} corresponds to the explanation based on that specific post. The last column, \say{Relevant Explanation}, is where you have to fill in 0 or 1.\\
	
	To measure the assessment effort, we ask you to record the time spent on fully evaluating the explanations presented.
}

{
	\onecolumn
	\section{Prompt for conversational LLMs}
	\label{appendix:prompt}
	
	\begin{figure}[ht]
		\centering
		\includegraphics[width=1\linewidth]{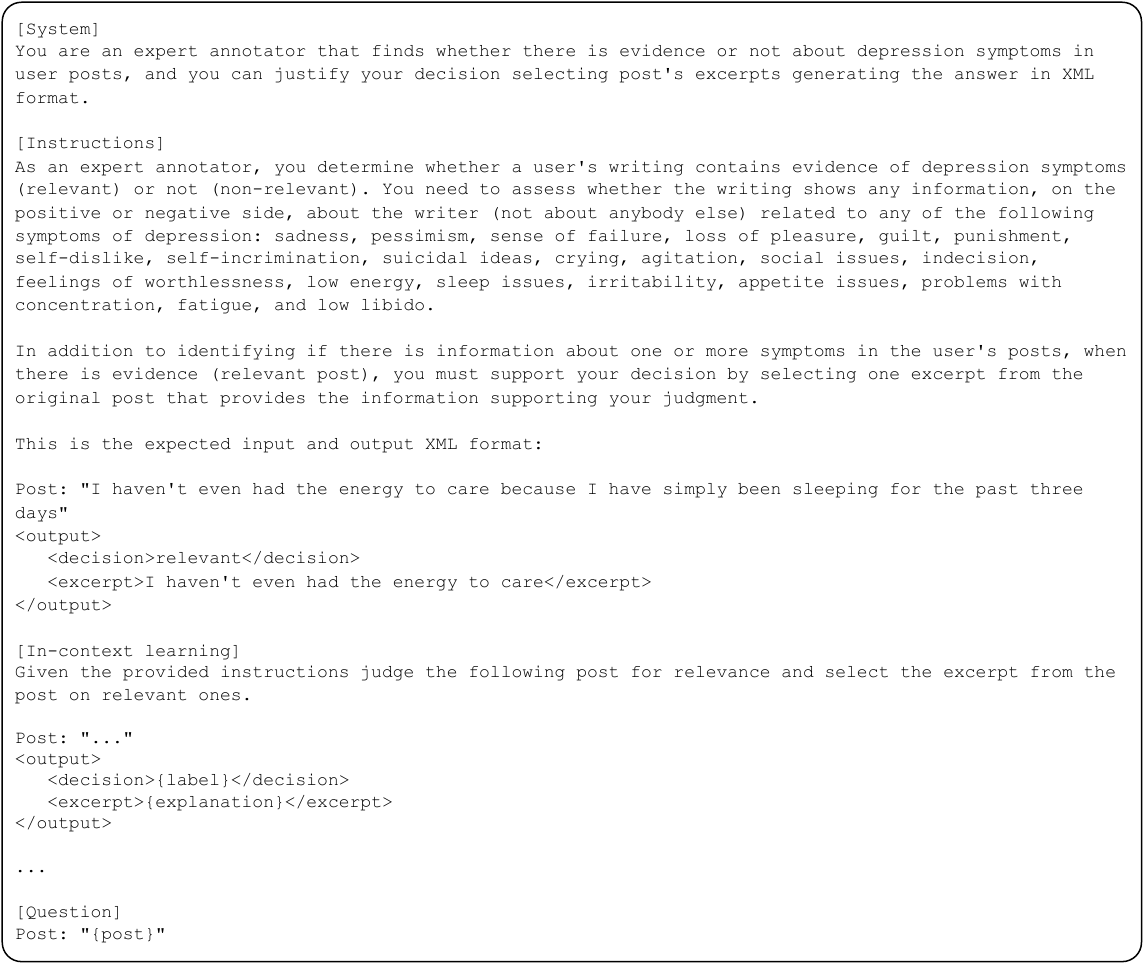}
		\caption{Prompt used for our experiments with conversational LLMs.}
		\label{fig:prompt}
	\end{figure}
}

\end{appendices}

\pagebreak



\end{document}